\documentclass[magazine,journal]{IEEEtran}
\usepackage{fancyhdr}
\usepackage{xspace}
\IEEEoverridecommandlockouts

\usepackage{lipsum} % For dummy text
\usepackage{url}    % For hyperlink in notice

\usepackage{graphicx} 
\usepackage{pifont}
\usepackage{amsmath}
\usepackage{amsfonts}
\usepackage{amssymb}  
\usepackage{amsfonts}
\usepackage{algorithmic}
\usepackage{array}
\usepackage{multicol}
\usepackage{amsmath}
\usepackage[inline]{enumitem}
\usepackage{textcomp}
\usepackage{stfloats}
\usepackage{url}
\usepackage{verbatim}
\usepackage{graphicx}
\usepackage{caption}
\usepackage{subcaption}
\usepackage[ruled, vlined, linesnumbered]{algorithm2e}
\usepackage{cite}
\usepackage{booktabs}
\usepackage{amssymb,amsfonts,bm}
\usepackage{amsmath,tikz}
\usepackage{color}
\usepackage{mathtools}
\usepackage{hyperref} 
\hypersetup{
    colorlinks=true,
    linkcolor=black,
    citecolor=black, % Set the cite color to black
    filecolor=magenta,      
    urlcolor=blue,
    }

\usepackage{rotating}
\usepackage{blkarray}
\usepackage{physics}
\usetikzlibrary{arrows}
\usepackage{makecell}

\usepackage{amsthm}
\usepackage{comment}
\usepackage{multirow}
\usepackage{geometry}
\graphicspath{{image/}}

\begin{document}

\title{\LARGE \bf
Natural Multimodal Fusion-Based Human–Robot Interaction: Application With Voice and Deictic Posture via Large Language Model\\

\author{Yuzhi Lai$^{1}$,~Shenghai Yuan$^{2*}$, ~\textit{Member,~IEEE}, ~Youssef Nassar$^{3}$, ~Mingyu Fan$^{4}$,\\ ~Atmaraaj Gopal$^{5}$, ~Arihiro Yorita$^{6}$, ~Naoyuki Kubota$^{7}$ and ~Matthias Rätsch$^{3}$}

% \thanks{
% $^*$ Corresponding Author.}
\thanks{
$^*$ Corresponding Author. This work is supported by a grant of the EFRE and MWK ProFö-R\&D program, no. FEIH\_ProT\_2517820 and MWK32-7535-30/10/2. This work is also supported by ``CAIpirinha - Conversational AI and Personalized Interaction for Risk-aware Navigation with Human Awareness'' Förderkennzeichen: BW7\_1030/02, Funding Program ``Invest BW - Innovation III''.
This work is additionally supported by the National Research Foundation, Singapore, under its Medium-Sized Center for Advanced Robotics Technology Innovation.}

\thanks
{$^{1}$University of Tuebingen,  Geschwister-Scholl-Platz, 72074 Germany.
   {\tt yuzhi.lai@uni-tuebingen.de}}%

\thanks{$^{2}$Nanyang Technological University, 50 Nanyang Avenue, Singapore 639798, 
   {\tt shyuan@ntu.edu.sg}}%

\thanks{$^{3}$University Reutlingen, Alteburgstraße 150, 72762 Germany.
   {\tt \{name.surname\}@reutlingen-university.de}}%

\thanks{$^{4}$Donghua University, 849 Zhongshan West Street, Shanghai 200051, 
   {\tt fanmingyu@dhu.edu.cn}}%
\thanks{$^{5}$Neura Robotics GmbH, 44 Gutenbergstraße, Metzingen 72555, 
  {\tt atmaraaj.gopal@neura-robotics.com}}%
\thanks{$^{6}$Kwansei Gakuin University, 1-155 Uegahara 1bancho, Hyogo 662-8501}
\thanks{$^{7}$Tokyo Metropolitan University, Tokyo, Hachioji, Minamiosawa, 1-ch.-1 }

}

\maketitle
\IEEEpubid{\begin{minipage}{\textwidth}\ \\[55pt]
\centering
\fbox{%
\parbox{\dimexpr\textwidth-2\fboxsep-2\fboxrule}{%
\centering
\footnotesize
This work has been accepted for publication in IEEE Robotics and Automation Magazine (RAM) © 2025 IEEE.\\
Personal use of this material is permitted. Permission from IEEE must be obtained for all other uses,\\
including reprinting/redistribution, creating new works, or reuse of any copyrighted components of this work in other media.
}%
}
\end{minipage}}

%\thispagestyle{empty}
%\pagestyle{empty}
%relocation tasks such as placing one object inside another
%\markboth{Preprint submitted to IEEE Robotics \& Automation Magazine}
%{How to Use the IEEEtran \LaTeX \ Templates}
%%%%%%%%%%%%%%%%%%%%%%%%%%%%%%%%%%%%%%%%%%%%%%%%%%%%%%%%%%%%%%%%%%%%%%%%%%%%%%%%
\begin{abstract}
Translating human intent into robot commands is crucial for the future of service robots in an aging society. Existing Human-Robot Interaction (HRI) systems relying on gestures or verbal commands are impractical for the elderly due to difficulties with complex syntax or sign language.  To address the challenge, this paper introduces a multimodal interaction framework that combines voice and deictic posture information to create a more natural HRI system. Visual cues are first processed by the object detection model to gain a global understanding of the environment, and then bounding boxes are estimated based on depth information. By using a large language model (LLM) with voice-to-text commands and temporally aligned selected bounding boxes, robot action sequences can be generated, while key control syntax constraints are applied to avoid potential LLM hallucination issues. The system is evaluated on real-world tasks with varying levels of complexity using a Universal Robots UR3e manipulator. Our method demonstrates significantly better HRI performance in terms of accuracy and robustness. To benefit the research community and the general public, we will make our code and design open-source. 
\end{abstract}
% \textit{Index Terms}: Human-robot Interaction, Intent recognition, multimodality perception, Large Language Models

%%%%%%%%%%%%%%%%%%%%%%%%%%%%%%%%%%%%%%%%%%%%%%%%%%%%%%%%%%%%%%%%%%%%%%%%%%%%%%%%

\section{INTRODUCTION}
\begin{figure*}[thpb]
      \centering
      %\vspace{-15pt}
      \includegraphics[width=1\textwidth]{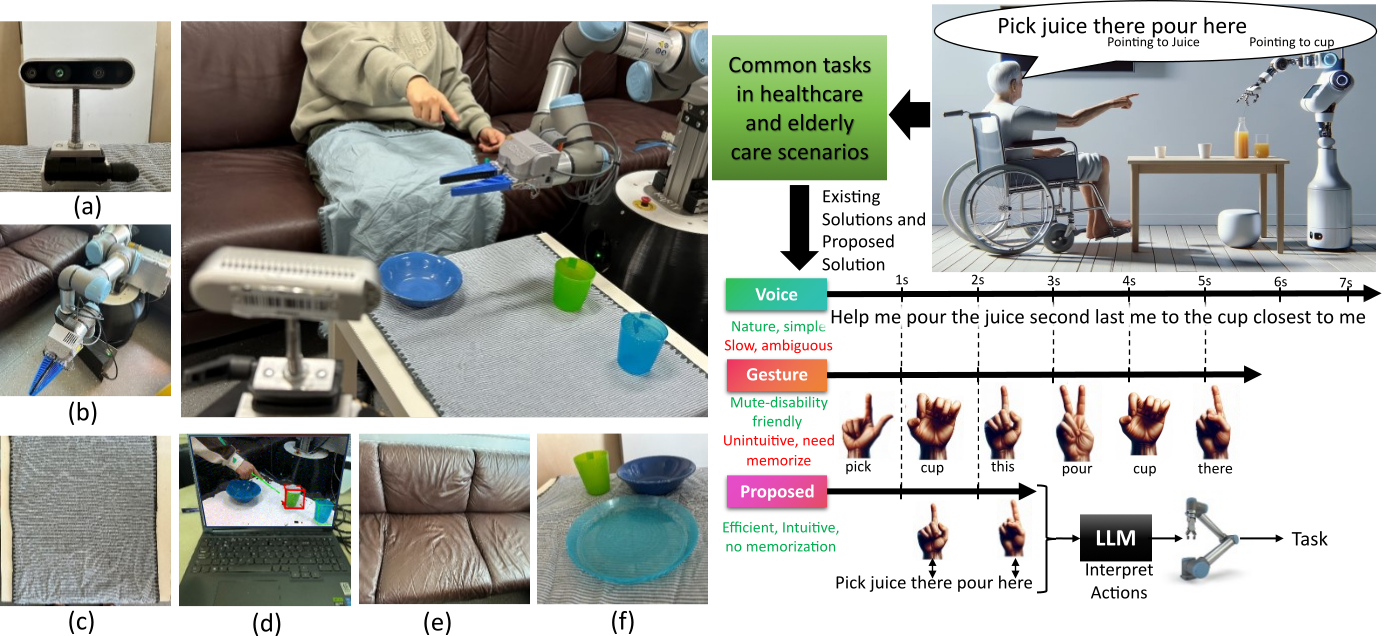}

      \caption{ Proposed voice-posture fusion HRI method has superior efficiency and requires no memorization of key syntax, which is ideal for elderly and healthcare applications. (a) Depth camera, (b) Robot manipulator, (c) Robot operating space, (d) Visual feedback, (e) User space, (f) Objects for experiment.}
      \label{figurelabel}
 
\end{figure*} % todo 图上写一下优点 

With an aging population, the demand for efficient caregiving solutions is growing, yet labor costs continue to increase. Robotics manipulators \cite{qi2024air} have shown promise in alleviating these challenges by automating tasks \cite{liu2025handle} that traditionally require human intervention \cite{9335006,li2024jacquard}. However, one of the most critical and unresolved issues is ensuring effective Human-Robot Interaction (HRI) \cite{weber2020distilling} for elderly users. For HRI systems to truly benefit this population, they must offer interaction methods that are intuitive and natural, resembling everyday human communication. Current systems often rely on humans to memorize complex language syntax or master complex hand gestures \cite{hanggesture,leapmotion}, which are impractical for the elderly. This highlights the urgent need for a simpler \cite{lai2024intuitive} yet highly effective method that allows robots to understand and execute commands from elderly users with ease and reliability.
 
In the past year, large language models (LLMs) \cite{zhang2023large,zhao2023chat} have emerged as promising tools for HRI \cite{10341989}. Their advanced reasoning and language capabilities make them promising for improving communication between humans and robots. However, directly applying LLMs to HRI presents several challenges. First, LLMs often require users to input detailed and structured text commands, which can be tedious and difficult to understand. Second, without integrated sensing capabilities, LLMs struggle to comprehend the environmental context \cite{yang2024fast,10715566} or specific actions, limiting their effectiveness in real-world applications. Finally, LLMs are prone to hallucinations, generating inaccurate or unsafe responses, which can lead to harmful outcomes when used in control systems without close monitoring. These challenges highlight the need for a more robust integration of LLMs into HRI systems.

To address the challenges, we introduce an \textbf{N}atural \textbf{M}ulti-\textbf{M}odal fusion-based \textbf{HRI} framework (NMM-HRI) that recognizes voice and posture, enabling users to convey their intentions to robots. We use simple and intuitive verbal language to compile sets of actions, while deictic postures identify objects or locations for interactions. \textcolor{black}{By combining voice commands with deictic postures, our approach resolves ambiguities in language-based systems, reducing cognitive load in gesture-based systems and providing a more intuitive and natural interaction experience.} Additionally, we incorporate an LLM to compile these actions and goals, generating robot action sequences. \textcolor{black}{Unlike rule-based systems or simpler models, LLMs leverage their extensive reasoning ability to handle complex contextual understanding and generate the most reasonable action sequences.} The generated action sequences undergo further adjustment of the structure of the output response to ensure structural consistency for safety purposes. Our method allows for the efficient construction of complex sequences of control actions, surpassing the speed of previous benchmarks by almost 50\%.
Our main contributions are summarized below:

\begin{itemize}
	\item 
We propose NMM-HRI, a parallel multimodal HRI method for robot manipulation. It efficiently constructs complex temporal control sequences using simple parallel inputs, processed by LLM to generate feasible actions.
	\item 
Our proposed system seamlessly generates robot control sequences through language, posture, and environmental input. This is achieved by structuring the output response tokens to mitigate LLM hallucination issues in the HRI setting, ensuring safety.
	\item We benchmark our system against state-of-the-art HRI methods, showcasing strong performance with minimal syntax token memorization and rapid input speed. For the benefit of the community, our system design, algorithms and solutions will be open-source at \url{https://github.com/laiyuzhi/NMM-HRI}.%\href{https://github.com/laiyuzhi/multimodal-HRI-toolbox}{Github}.
\end{itemize}

%The source code, accompanying video, and additional materials are available on the project webpage
%http://xxxxxxxxxxxxxxxxxxxxxxxxxxxxxxxxxxxxxxxxx.html.

\section{RELATED WORK}
As perception and navigation solutions advance \cite{yuan2021survey}, robots are increasingly integrated into daily life. However, robots designed for complex tasks, such as surgical \cite{7924395} and rehabilitation applications \cite{8684289}, face stringent safety requirements and must be validated for many years before widespread use in real-world scenarios. In contrast, service robots, particularly for elderly and patient care, demonstrate greater immediate potential. However, current service robot systems often rely on single-modal inputs, such as pure hand gestures \cite{hanggesture}, voice commands \cite{stepputtis2020language}, or combinations of body language, such as hand gestures and body postures \cite{c10}. Hand gesture-based HRI methods \cite{hanggesture}, for example, frequently use the Leap Motion sensor to detect hand movements and translate them into commands \cite{c14}. However, these sensors have a limited field of view, restricting their applicability to broader environments. To address this, some researchers \cite{c10} have introduced Kinect cameras to extend the range of the sensor and detect both human posture and gestures. Nonetheless, for elderly users, accurately memorizing and executing the required gestures for commands remains a significant challenge, limiting the effectiveness of these approaches.
Voice commands have always been a natural candidate for HRI problems. People often employ various approaches, such as LSTM, hidden Markov models, or an attention-based encoder-decoder network \cite{asfour2019armar}, to parse voices into actionable commands. One good example is the Amazon Alexa AI assistant, which can turn on and off lights with voice commands. However, controlling the voice-activated robot with a higher degree of freedom presents several challenges, including the ambiguity in describing the scene \cite{esfahani2020unsupervised} and the potential for sentences to have multiple interpretations.

multimodal approaches have great potential in generating accurate HRI responses. 
Early approaches \cite{8360084} only allow simple command syntax, which dramatically limits its applications. 
In recent years, wearable mixed reality (MR) devices \cite{groechel2022tool}, such as the Apple Vision Pro, have demonstrated their capabilities as HRI tools. They often come equipped with multimodal inputs like gaming joysticks, gaze detection, head orientation estimation, and voice commands. However, these methods have several issues, including:
(1) MR takes a considerable amount of time for humans to learn, with a steep learning curve; 
(2) Gaze and head-orientation-based HRI often require individualized calibration solutions to infer intentions with reasonable accuracy, thus introducing more problems.
(3) The devices tend to be excessively heavy, posing a challenge for the average user, let alone individuals who are elderly or unwell.
(4) They can induce feelings of dizziness and nausea.
For elderly individuals or those in need of medical attention, the use of MR devices for cyber-retirement is unlikely to be well-received. \\
Several other methods have incorporated multimodal inputs as redundant systems to enhance fail-safety. Although some claim to offer universal approaches for multimodality in HRI \cite{c15}, these are often limited to late fusion frameworks that only partially address failure scenarios. Other modalities, such as electromyography, facial expressions, and voice signals \cite{c9}, have been combined to control devices like wheelchairs through redundancy. However, these combinations are often impractical due to the variation in electromyographic signals and facial expressions between users, making it impossible to generalize their use effectively.

Existing multimodal approaches have yet to meet expectations for general use, particularly for elderly users or patients who struggle with complex sign or command languages. A major challenge is the ineffective fusion of different information sources. Recent works on Visual Language Models (VLMs) \cite{ConstantinEYBW22} have been proposed to address this issue in HRI. However, these models are often application-specific or tailored to specific environments, lack comprehensive benchmarks against traditional methods, and require substantial GPU resources for processing. In contrast, large language models (LLMs) offer strong reasoning and emergent capabilities while maintaining reasonable computational requirements, making them promising candidates for effectively fusing and processing multimodal information \cite{zhao2024adaptive}.

Most research in robot action generation involves predefined domain searches, unconstrained exploration \cite{jin2024robust}, behavior trees, and Bayesian inference. In one study, the authors integrated LLMs into robot action generation, defining a pipeline that converts human intentions into robotic action sequences using prompts and task-relevant APIs \cite{c7}. This approach provides a more intuitive and convenient method for user interaction and robot control \cite{li2025ua}. However, LLMs cannot directly obtain information from sensors \cite{liang2025unsupervised}, and Visual Language Models (VLMs) often require significant computational resources \cite{ConstantinEYBW22}. Furthermore, since LLMs are prone to hallucination, combining LLM with VLM increases the likelihood of errors and mistakes, often necessitating multiple trials for successful execution.

\vspace{-10pt}
\section{PROBLEM DEFINITION}
\label{sec:PROBLEM FORMULATION}
%This section provides a mathematical description of our framework for inferring human intention.
\textcolor{black}{
The goal of our proposed solution is to derive a parallel multimodal command sequence, which is then translated into robotic action sequences through the use of an LLM. To achieve that, we need to define the problem and the set of mathematical representations. }
% \textcolor{black}{blue}
%%\vspace{-9pt}The $q(.) \in \mathbb{R}^6
%$ represents the robotic arm manipulator joint state space information controlled by robot action sequences  $\mathbb{A}$. 

\vspace{-10pt}
\subsection{Problem Formulation}
\textcolor{black}{
Let $\Xi(.)$ denote the object prior information represented by object class $\kappa$  and object \textcolor{black}{representation $\mathcal{O} \in \mathbb{R}^5$ including 3D position $\xi \in \mathbb{R}^3$, height $h \in \mathbb{R}$  and width $b \in \mathbb{R}$.} The $q(t) \in \mathbb{R}^7
$ represents the state of the robot end-effector at time $t$, including the 6D pose and the opening angle of the gripper. 
The prior observation tuple of the environment $\mathcal{S}$ can be constructed by $\mathcal{S}=(\Xi(\kappa, \textcolor{black}{\mathcal{O}}),\ q(t))$. }

\textcolor{black}{
To control the manipulator $q(t)$ under constraint $\mathcal{S}$, it is essential to infer complete human intention $I$ from a set of sparse keywords $\mathcal{C}$, which consist of object references $I_\mathbb{O}$ and action references $I_\mathbb{A}$.
The NMM-HRI system uses audio and RGBD sensors to generate time series verbal instruction sets $\mathcal{V}$ and human postures $\mathcal{B}$. Object references $I_\mathbb{O}$ specify the target object to be interacted with according to $\mathcal{B}$, while action references $I_\mathbb{A}$ indicate type of action to be performed with object based on verbal command $\mathcal{V}$.
}

Therefore, the action intention can be considered as a mapping function from verbal language $\mathcal{V}$ to action intention $I_\mathbb{A}$, defined by $\mathcal{M}: I_\mathbb{A} = \mathcal{M}(\mathcal{V})$.
Similarly, the object intention $I_\mathbb{O}$ can be represented by the posture reference $\mathcal{B}$, which interacts with environmental observations $\mathcal{S}$. This representation is defined by the mapping function $\mathcal{P}$, where $I_\mathbb{O} = \mathcal{P}(\mathcal{B}, \mathcal{S})$.
%Therefore, we have know the mapping $\mathcal{M}: I_a=\mathcal{M}(\mathcal{V})$ because we use verbal language to represent action intention. And also we have to know mapping $\mathcal{P}: I_o=\mathcal{P}(\mathcal{B},\ \mathcal{S})$. We use deictic posture from human skeletion to identify object intention in the scene.

%\textit{Problem definition}: The multimodal observed result $o$ is processed to derive human intention $i$, which is subsequently transformed into a robotic action sequence $a$.

%The \textit{multimodal observed result}  $o$ is defined as $o = [v, p, s]$. This encompasses time-series verbal features $v$ and deictic postures $p$. The variable $s$ encapsulates the context of the scene, detailing the status of both objects and the robot  (robot end-effector, position of the recognized objects, etc.).

%The \textit{Human intention} modification of the scene state $s$. For predicting human intentions, we categorize them into two components: object intention and action intention. \textit{Object Intention} specifies the target object for manipulation by humans. \textit{Action Intention} denotes the specific action that humans wish to perform on the object defined by the object intention.

\vspace{-10pt}
\subsection{Parallel multimodal Command Sequence}
%Individual verbal features are capable of conveying diverse information, including the desired action and the class of the desired object. Deictic posture is capable of target object.
%Action intention and object intention can be obtained through parallel multimodal command sequence and scene. Here we list the components of multimodal command sequence.
%For more information please refer to Sec. \ref{section:MATERIALS AND METHODS}.  
Action and object intentions are derived from parallel multimodal command sequences and scenes. Below, we outline the components of the multimodal command sequence.
\begin{itemize}
 
\item \textbf{Verbal class command:} This command defines the specific class relevant to object intention. An object will only be detected if its class $\kappa$ corresponds to verbal class command.
%, such as \textit{bowls}, \textit{bottles}, \textit{cups}, and \textit{plates}. 
%When numerous objects of various categories are present together, specifying and detecting a unique class of object can significantly enhance the accuracy of determining the object intention.

\item \textbf{Verbal action command:} This command is assigned to various intended actions. Two distinct actions can be combined to construct a complex temporal movement, such as first picking up a cup and then pouring water into a bowl.

\item \textbf{Verbal pronoun command:} This command, such as \textit{this}, \textit{there} or \textit{that}, is employed in conjunction with deictic posture.
When the demonstrative pronoun is recognized by the system, it records the location of the object that has been selected through the deictic posture.

\item \textbf{Verbal metric command:} This optional command enhances input verbal information. This command could include variables such as the angle of inclination for a pouring action or the speed of various actions.

\item \textbf{Deictic posture:} This specific posture is instrumental in aiding users to select an object within a scene as the object intention. By human skeleton detection, we obtain the deictic posture, $r$, representing the direction of the user's right forearm. The distance of an object  $\Xi_i \in \Xi$ in the scene $\mathcal{S}$ with 3D position $\xi_i$ to the vector $r$ is outlined as $d_i(r, \xi)$:
\vspace{-5pt}
\begin{equation}
d_i =\sqrt{\frac{\left | (r_2-r_1)\times (r_1-\xi_i) \right |^2 }{\left | r_2-r_1 \right |^2 } }, \label{eq:dis}
\end{equation}
where $r_1$ and $r_2$ are two random points from $r$. We specify the object closest to the deictic posture as the object intention, so the mapping $\mathcal{P}$ can be rewritten as:

 \begin{align}
     I_\mathbb{O} &= \mathcal{P}(\mathcal{B}, \ \mathcal{S}) \notag , \\
     &\triangleq \mathcal{P}(r,\ \xi) . \label{eq:dis1}
 \end{align}

%The term $S_{poses}$ refers to the 3D positions of detected objects within the scene. Additionally, $o_{i}^{dist}$ quantifies the closest distance from the detected object to the vector $p_{line}$, as detailed:
    
\end{itemize} 
Mapping $\mathcal{M}$ converts all inputs from verbal language into text and completes a query task to find the action intention $I_\mathbb{A}$ corresponding to the verbal action command.
%In more detail, by adjusting the multimodal command sequence $\mathcal{C}$ we can obtain human intention $I$ consisting of object intent $I_\mathbb{O}$, action intention $I_a$ and metric parameter $\omega$:
Through adjustment of the multimodal command sequence $\mathcal{C}$, we can derive the human intention $I$, which encompasses the object intent $I_\mathbb{O}$, action intention $I_\mathbb{A}$, and metric parameter $\omega$.
\begin{align}
    I &\triangleq \mathcal{F}(\mathcal{C})  \notag ,\\
      &= \mathcal{F}(I_\mathbb{A}, \ I_\mathbb{O}, \ \omega ) .\label{eq:multimodal command} 
\end{align}

\begin{figure}[t]
      \centering
      \vspace{-10pt}
      \includegraphics[width=0.48\textwidth]{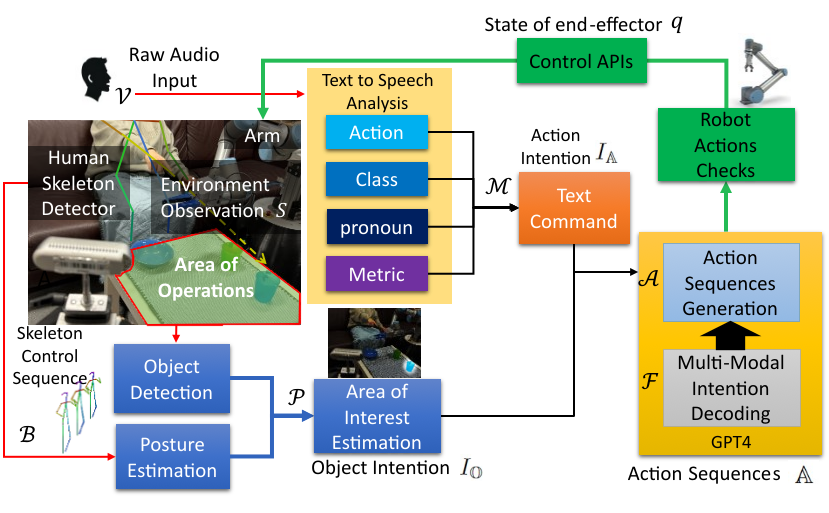}
      \vspace{-10pt}
      \caption{ \textcolor{black}{ System Overview.
$\mathcal{V}$ represents voice command, $\mathcal{B}$ represents human posture, $\mathcal{M}$ is mapping verbal features to action intention $I_{\mathbb{A}}$, $\mathcal{P}$ is mapping human posture and environment observation $\mathcal{S}$ into object intention $I_{\mathbb{O}}$. 
GPT4 decodes the multimodal commands and generates the action sequences $\mathbb{A}$. Finally, the state of end-effector $q$ is changed by the control APIs.}  }
      \label{systemdiagram}
      \vspace{-29pt}
\end{figure}

The function $\mathcal{F}$ represents the encoding process from a multimodal command sequence to human intention, as shown in Fig. \ref{systemdiagram}. \textcolor{black}{
%The input process of multimodal command will be in detail later. 
In our work, GPT4 \cite{c23} is utilized to decode command sequence $\mathcal{C}$ into intention $I$. } The robotic action sequence $\mathbb{A}$ is derived from intention $I$ with mapping $\mathcal{A}: \mathbb{A}=\mathcal{A}(I)$. The mapping is also accomplished using GPT4. \textcolor{black}{Finally, GPT4 uses the action sequences $\mathbb{A}$ passed through the check to control the state of the end-effector $q(t)$ to fulfill the human intention $I$.}  %The overall structure of this approach is depicted in Fig. \ref{systemdiagram}

   \vspace{-7pt}
\subsection{Construction of Complex Command Sequence}

Simple robotic actions require only the specification of the action type, without any additional parameters or object dependencies. An example of this is \textit{go initial position}. For actions like \textit{Pick up a cup}, it is necessary to specify both action and object intention. 
To achieve more complex temporal control, such as \textit{picking up a cup and then tilting it at a 90-degree angle to pour into a bowl}, our multimodal command sequence allows for the construction of two subordinate commands. In this way, the multimodal command sequence becomes:
 %\vspace{-2pt}
\begin{equation}
    \textcolor{black}{\mathcal{C}  = \left \{ \mathcal{C}_1, \mathcal{C}_2 \right \} =\left \{ I_{\mathbb{A}1},\ I_{\mathbb{O}1}, \ I_{\mathbb{A}2},\ I_{\mathbb{O}2},\ \omega \right \}} , \label{eq:multimodal command2} 
\end{equation}
 %\vspace{-2pt}
where, $ \mathcal{C}_1$ and $\mathcal{C}_2$ represent the respective subordinate commands, with each subordinate command encompassing its own action intention $I_\mathbb{A}$ and object intention $I_\mathbb{O}$.
%The complexity of a multimodal command sequence is contingent upon the intricacy of the robot's actions. A multimodal command sequence necessitates only one essential parameter $a_i$. We define the complexity of a multimodal command sentence $C$ as follows:
%\begin{equation}
%    C_c= len(o_i) + len(mp)
%\end{equation}
%Simple robotic actions require solely the specification of the action type, without any additional parameters or object dependencies. An example of this is \textit{go initial position}, based on a grid world step. For such actions, the multimodal control sequence complexity ($C_c$) is 0. For actions like \textit{Pick up a cup}, it is necessary to specify both action and object intention, resulting in a multimodal control sequence complexity of $C_c=1$. More complex robotic actions may depend on combinations with two subordinate control sequence in the scene, such as  \textit{Pick up a cup and pour the water in the bowl with 90 degree}. $N_1$ and $N_2$ represent the complexities of each subordinate control sequence, respectively: $N_1=1$, $N_2=2$ and $C_c = 3$
Our system contains the following set of robotic actions
with increasing complexity:
\begin{itemize}
    \item Without object dependency: go initial position, throw, \textcolor{black}{flush, etc.}
    \item  With object dependency: pick, put, pour, \textcolor{black}{clean, push, etc}.
    \item  Combination of subordinate command sequences: pick + put, pick + pour, pick + throw, pick + go initial, etc.
\end{itemize}
%\vspace{-5pt}
\section{methodology}
\label{section:MATERIALS AND METHODS}
%\vspace{-5pt}
\textcolor{black}{
Fig. \ref{systemdiagram} illustrates the overall system, designed for indoor healthcare or elderly care scenarios where the robot performs tasks based on human input. To enable concise and intuitive interactions, the system integrates several subsystems, including speech-to-text conversion, object detection, posture detection, and action sequence generation and execution. These components work together to ensure that the robot can accurately interpret and carry out the user's commands.}
%The overall system flowchart is shown in Fig. \ref{systemdiagram}. We assume it is an indoor healthcare or elderly care scenario where the robot must fulfill tasks given by the human. The user interactions need to be concise and intuitive. In order to achieve this goal, several subsystems need to be constructed, including speed-to-text conversion, object detection, posture detection, action sequence generation, and execution. %Visual inputs are collected through the Intel Realsense D435i RGBD camera, while auditory signals are captured through a USB microphone. 
%\vspace{-8pt}

\vspace{-10pt}
\subsection{Speech-to-Text Conversion}
\textcolor{black}{
To understand human verbal commands, a speech-to-text module is essential. This module converts the raw audio input into meaningful elements such as class references, pronoun references, intended actions, and metric parameters. Unlike traditional systems that process complete sentences, our approach primarily handles fragmented commands, often consisting of partial words combined with posture cues. This imposes specific constraints on the tools used. After comparing various methods \cite{c24}, we selected the VOSK \cite{c19} for its ability to process partial input and distinguish between different speakers. }
% people may shoot 

%In work~\cite{c18}, after comparing five of the most popular real-time speech-to-text tools, the VOSK API~\cite{c19} and WhisperAI~\cite{c20} are particularly well-suited for integration with robotic systems.
%In our system, we have chosen to employ the VOSK API, primarily due to its capability to recognize individual speakers.

 %\vspace{-8pt}
 \vspace{-10pt}
\subsection{Object Detection}
\textcolor{black}{
The object detection model extracts both bounding boxes and object classes using visual cues. We evaluated a few models such as YOLOv5 \cite{redmon2016you}, YOLOv6, YOLOv8, and YOLO-World \cite{cheng2024yolo}, selecting YOLO-World for its real-time open vocabulary detection capabilities. Our results show that YOLO-World accurately recognizes everyday objects (i.e., shampoo, mug, bottle, scissors) with high reliability. The 2D position of the object within the RGB image is represented by the center of its bounding box. Following object detection, a coordinate transformation step uses the depth map aligned with the RGB image and the 2D object information to determine the object's 3D representation, $\mathcal{R}$, in the camera frame. The 3D position $\xi$ is then used to calculate the distance, as described in Eq. \ref{eq:dis}. The width of the object $b$ is used to calculate the opening angle of the gripper. The height of the object is used to calculate the collision-free trajectory. }

\subsection{Deictic Posture Detection}

\begin{figure}[h]
      \centering
      \includegraphics[width=0.50\textwidth]{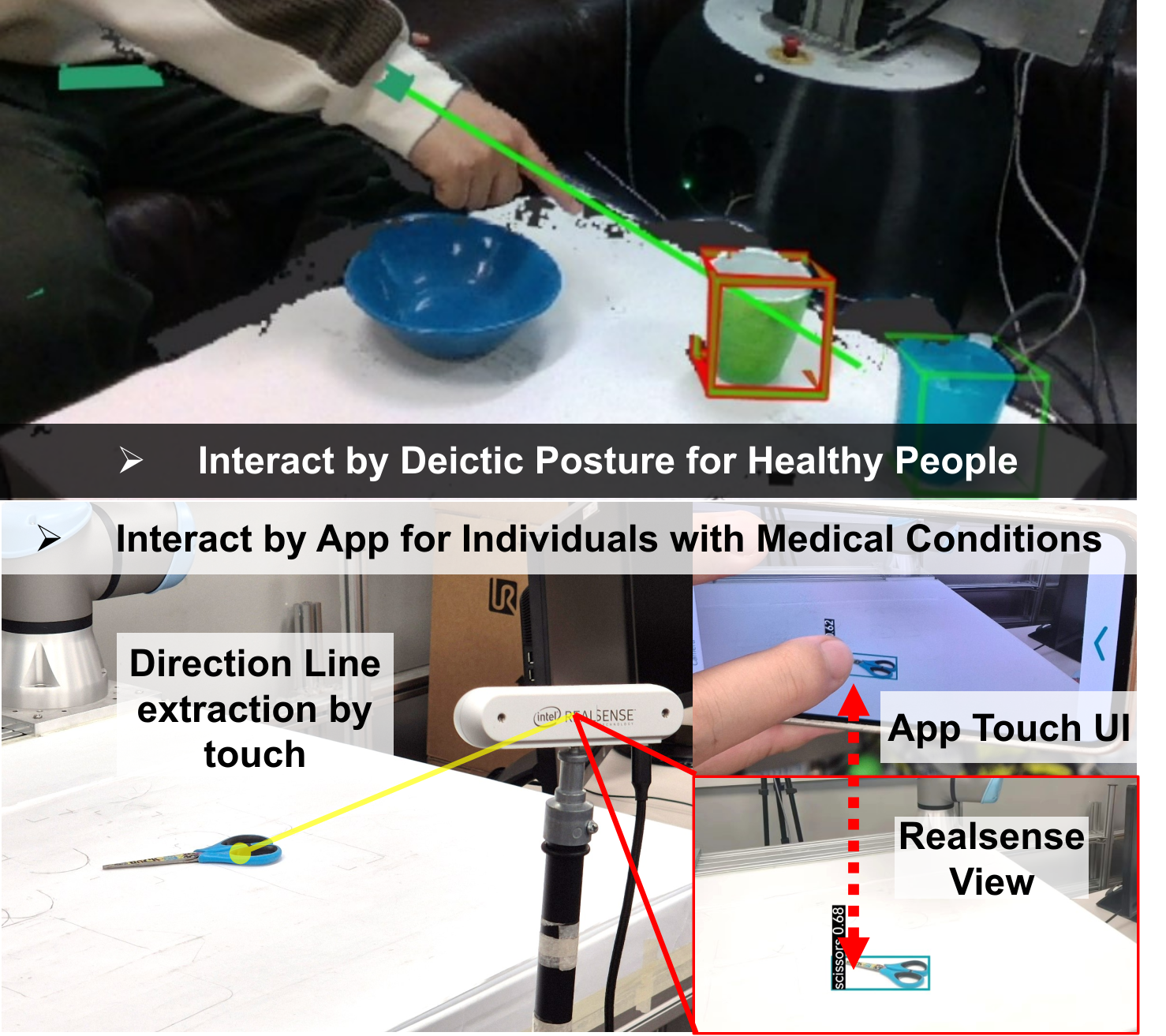}
      \caption{ Alternative ways of finding object reference.}
      \label{touch}
\end{figure}

The deictic posture represents a distinct type of static gesture used for inferring object reference, as referenced in Eq. \ref{eq:dis1} and Eq. \ref{eq:dis} % (refer to Equation \ref{eq:dis1} and Equation \ref{eq:dis} for how the target object is identified). 
The system captures the upper body of the human using an RGBD camera placed at an appropriate distance. Subsequently, the 2D human skeletons are tracked from the RGB image using OpenPose \cite{c12}. Then, the 3D human skeletons $\mathcal{B}$ are estimated within the camera frame based on the aligned depth map. We define the intention/direction line $r$ of the right forearm as the deictic posture. For cases where individuals cannot move their arms, an additional mobile App connects with RealSense cameras over a local area network to obtain the reference direction directly from camera view, as shown in Fig. \ref{touch}.

\vspace{-8pt}
The object reference can be calculated in each frame, but its information is only available when the user points to a detected object, the category of which is defined with a verbal class command and a verbal pronoun command is spoken. The accuracy of the deictic posture is evaluated in a separate experiment later.

  \vspace{-5pt}
 \begin{figure}[h]
      \centering
      \includegraphics[width=0.5\textwidth]{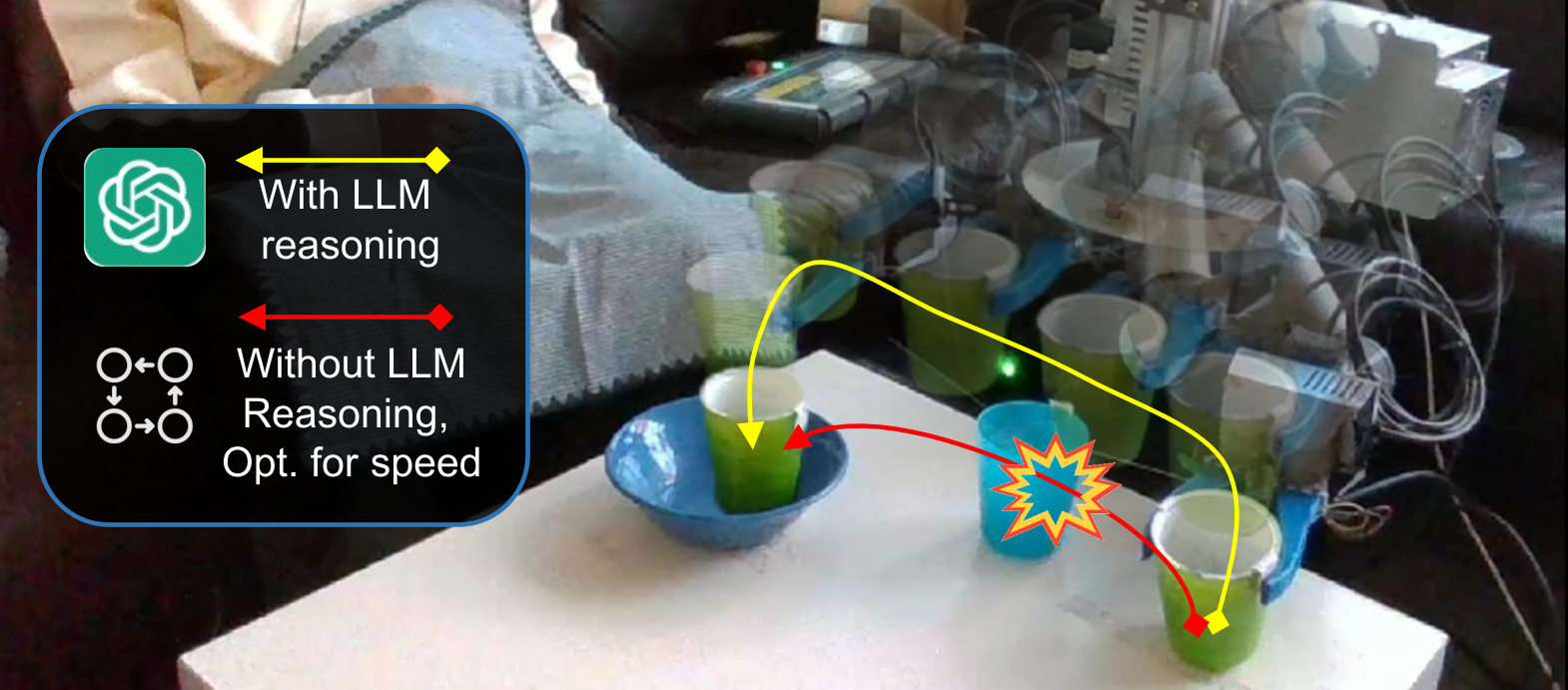}
      \vspace{-5pt}
      \caption{ \textcolor{black}{Collision-free trajectory generation.}}
      \label{pickobstacle}
      \vspace{-25pt}
   \end{figure} 

\subsection{Action Sequences Generation and Execution}
\begin{figure*}[thpb]
      \centering
      \includegraphics[width=1\textwidth]{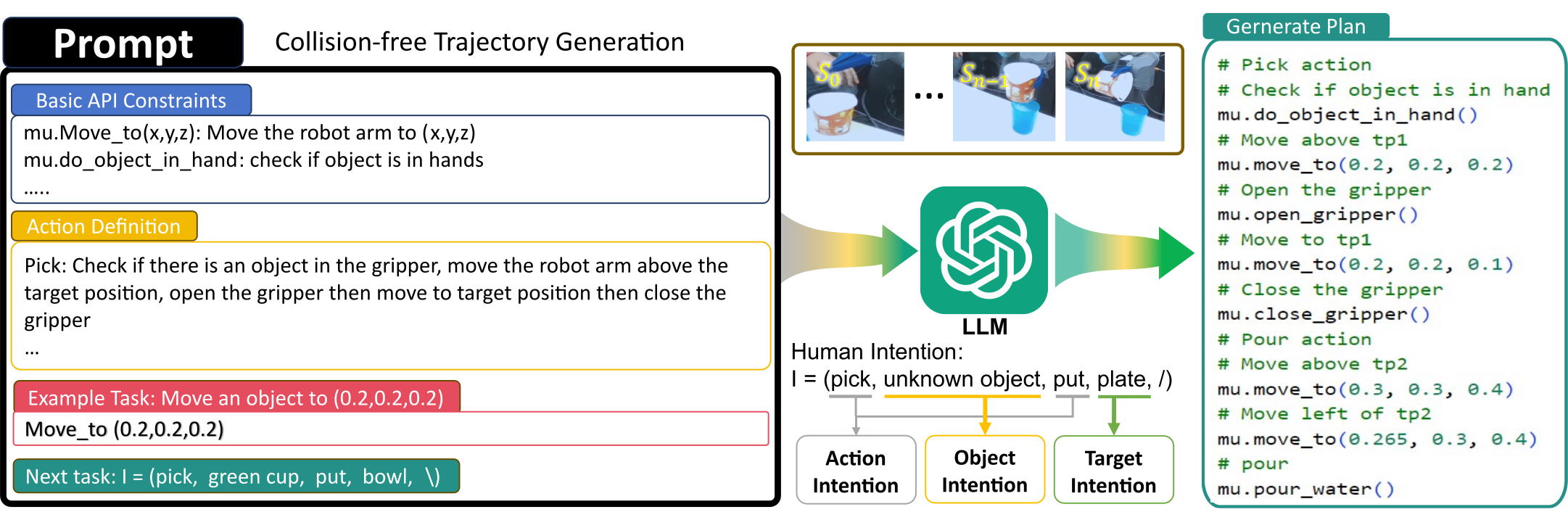}
      \caption{ The prompt is segmented into three sections: basic API constraints, action definition, and example tasks.}
      \label{actionsequence}
\end{figure*}

The human intention $I$ is encoded into a robot action sequences $\mathbb{A}$ through the mapping $\mathcal{A}$. Task planning and action sequence generation often require extensive domain knowledge about the state of robots and constraints. To streamline this process, we employ GPT4 to encode this high-level policy. We establish constraints on the output response tokens of the LLM by crafting prompts and setting constraints. This approach helps minimize the issue of hallucination in LLM. As illustrated in Fig. \ref{actionsequence}, the prompt consists of three parts:
\begin{enumerate}
    \item  Basic API Constraints: outlines the functionalities of APIs that can be utilized in task planning. LLM can only utilize these APIs and basic Python libraries such as numpy to build robot action sequences.
    \item  Action Definition: delineates the execution methodology for each action. Non-technical users have the flexibility to define new functionalities or modify the execution of each action in prompts using natural language. 
    \item Example Tasks: demonstrates how similar tasks are executed and guides the strategies for task planning using LLM. With this constraint, LLM will imitate this example task.
\end{enumerate}

%Additionally, we have incorporated state feedback from the environment, necessitating that the robot ascertain whether an object is already in its grasp prior to executing a pick task. Upon processing the prompt and human intentions, LLM generate corresponding action sequences and code.
Furthermore, we have integrated state feedback from the environment, requiring the robot to determine whether an object is already within its grasp before performing a pick task. Once the prompt and human intentions are processed, the LLM generates the corresponding action sequences and code. \textcolor{black}{ The generated action sequence is verified by LLM to ensure collision-free, as shown in Fig. \ref{pickobstacle}.} \textcolor{black}{For example, in the water pouring task illustrated in Fig. \ref{actionsequence}, LLM first understands the multimodal commands and then generates an action sequence based on the gripper state, as well as the height and position of the object. These sequences are restricted to actions defined by the basic API constraints and the action definition.}

\vspace{-10pt}
\subsection{Human-Robot Interaction}

\textcolor{black}{
Human-robot interaction in our system is enabled through parallel multimodal command sequences, where the user conveys intentions via verbal commands and deictic postures processed by $\mathcal{F}(.)$, with visual feedback (e.g., detected posture, selected object) provided by the graphical user interface.
}
The process of encoding a multimodal command sequence into human intention involves the following steps:

\begin{enumerate}
    \item The user begins by encoding the action intention $I_\mathbb{A}$ through a verbal action command. If this action requires object dependency, the sequence continues; otherwise, it is terminated with the \textit{finish} command.
    
    \item For object-dependent actions, the user encodes the class of the object (verbal class command) followed by a demonstrative pronoun (verbal pronoun command).

    \item Simultaneously, the system encodes the target object using deictic posture $r$ and object position $\xi$, computed by Eq. \ref{eq:dis}. The object selected during the pronoun command becomes the object intention $I_\mathbb{O}$.
    
    \item The metric parameters $\omega$, such as the pour angle, can be encoded verbally by the user to refine the action intention.

    \item Once the multimodal sequence is fully encoded, the user can either proceed with further subordinate commands or conclude the process with the \textit{finish} command, resulting in the final encoded human intention.
\end{enumerate}

\section{EXPERIMENTAL SETUP}

\subsection{Perception and Manipulator Setup}
\label{environment}

Visual inputs are collected through the Intel Realsense D435i RGBD camera, while auditory signals are captured through a USB microphone. 
Our system has been trialed with individuals of varying ages in real-world environments, as illustrated in Fig. \ref{figurelabel}. The scene contains a Universal Robots UR3e robot manipulator and several manipulation objects.
An Intel Realsense D435i RGBD camera opposing the robot is used for object detection and deictic posture detection. A laptop with a Nvidia 4060 GPU was used to process the multimodal data and show feedback images in Rviz.

\subsection{Description of Experimental Scenarios}
We designed a set of typical manipulation experiments of the increasing human intention complexity \textcolor{black}{according to eq. \ref{eq:multimodal command} and eq. \ref{eq:multimodal command2} with $I = \mathcal{F}(I_{\mathbb{A}1},\ I_{\mathbb{O}1}, \ I_{\mathbb{A}2},\ I_{\mathbb{O}2},\ \omega)$:} 
\begin{enumerate}  % C = (action,\ class,\ pronouns,\ x, \ metric)
\item[$\bullet$]$(home,\ -,\ -,\ -, \ -)$, $(throw,\ -,\ -,\ -, \ -)$
\item[$\bullet$]$(pick,\ cup,\ -,\ -, \ -)$
\item[$\bullet$] \textcolor{black}{$(push,\ plate,\ -,\ -, \ near)$} 
\item[$\bullet$] $(pick,\ cup,\ put,\ bowl,\ -)$, $(pick,\ cup,\ pour,\ cup,\ -)$
\item[$\bullet$] $(pick,\ cup,\ pour,\ bowl,\ ang = 90 ^{\circ} )$
\item[$\bullet$] \textbf{Multi-step tasks:} pick and throw rubbish, add water and pass, pour muesli and add milk.
\end{enumerate}
In our proposed method, we primarily evaluate the performance by measuring the user interaction time and success rate for each specific scenario. An additional goal is to evaluate the intuitiveness of the HRI system. The execution of all these tasks is depicted in the accompanying videos.

\section{RESULTS AND DISCUSSION}
\textcolor{black}{
We evaluated the performance of our system through a series of challenging experiments designed to assess criteria such as accuracy, user satisfaction, lighting robustness, and location robustness.}
% \label{RESULTS}
% We show results for the following aspects:
% \begin{enumerate*}
% \item Comparison of Efficiency and Intuitiveness;
% \item Comparison of Accuracy and Robustness;
% \item Online and field Survey on User Satisfaction;
% \item \textcolor{black}{Tests under adverse lighting condition;}
% \item \textcolor{black}{Extending tests in household tasks.}
% \item \textcolor{black}{Perception tests in diverse real-world environment.}
% \end{enumerate*}
% All the scenes and tasks are set so that the robot is able to complete tasks using objects in the scene. 
%The supplementary experiment will be uploaded to GitHub. 

\begin{figure}[thp]
\centering
\includegraphics[width=0.5\textwidth]{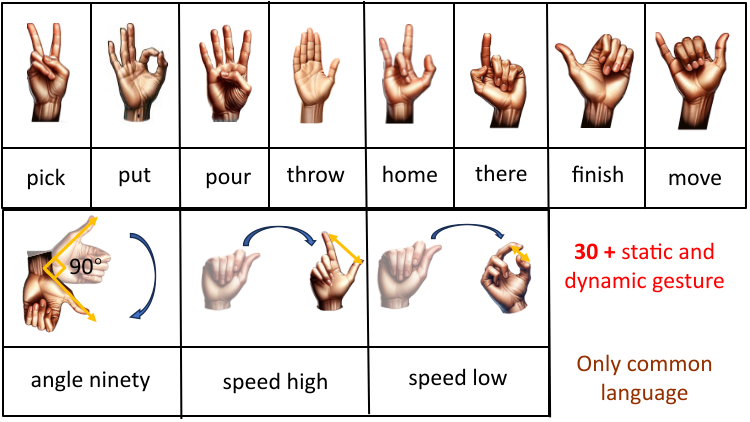}
\caption{  List of Gestures used in gesture-based HRI system \cite{hanggesture} and their corresponding verbal commands in our NMM-HRI experiments.} 
\label{mapping}
\end{figure}
\subsection{Baseline Selections} 
In our work, we compare our approach with other open-source SOTA unimodal HRI methods \cite{hanggesture, stepputtis2020language} and the multimodal method \cite{ConstantinEYBW22}. The baselines were selected primarily on the basis of similarity in action sequences and intent. The two unimodal benchmarks are gesture-based  \cite{hanggesture} and NLP-based  \cite{stepputtis2020language}, respectively. The multimodal approach is based on VLM \cite{ConstantinEYBW22}, where target objects \cite{yuan2014autonomous} are selected through dialogue.
\textcolor{black}{
The gesture-based HRI method \cite{hanggesture}, which utilizes the Leap Motion sensor, captures hand structure at specific localizations \cite{xu2025airslam,chen2025relative,yuan2024large,li2025helmetposer} within a narrow field of view.  The NLP and VLM approach \cite{stepputtis2020language, ConstantinEYBW22} demonstrated the use of language commands to direct the actions of a robot with a speech-to-text pipeline. \textcolor{black}{NLP-based methods employ word embeddings, attention mechanisms, and probabilistic reasoning to recognize objects described in natural language. In contrast, VLM-based methods utilize the CLIP model to resolve ambiguities in object selection and assist users in forming clearer expressions.} However, each baseline method has certain limitations and received numerous complaints from participants:
}

\textcolor{black}{
\begin{enumerate*}
\item The Leap Motion's limited field of view requires maintaining the hand consistently above the sensor, which is impractical for the elderly and patients.
\item Natural language commands often struggle to differentiate between similar objects using simple descriptors.
\item Gesture-based HRI requires memorizing a complex set of gestures, which grows increasingly complicated as the command set expands, as shown in Fig. \ref{mapping}. Those are the gesture commands that we ask the participant to memorize.
\item VLM-based methods require gestures to indicate the general direction of a target object, followed by an interactive dialogue to select it. This increases interaction time, and distinguishing between similar objects in the scene remains a challenge. 
\end{enumerate*}
}
%\vspace{-10pt}
  % \begin{figure}[thpb]
  %     \centering
  %     \includegraphics[width=0.4\textwidth]{image/compare1.png}
  %     \vspace{-5pt}
  %     \caption{\footnotesize Compared to other baseline HRI, our system required less time to input the same commands.}
  %     \label{episode}
  %     \vspace{-18pt}
  %  \end{figure}

  \begin{figure}[t]
      \centering
      \includegraphics[width=0.5\textwidth]{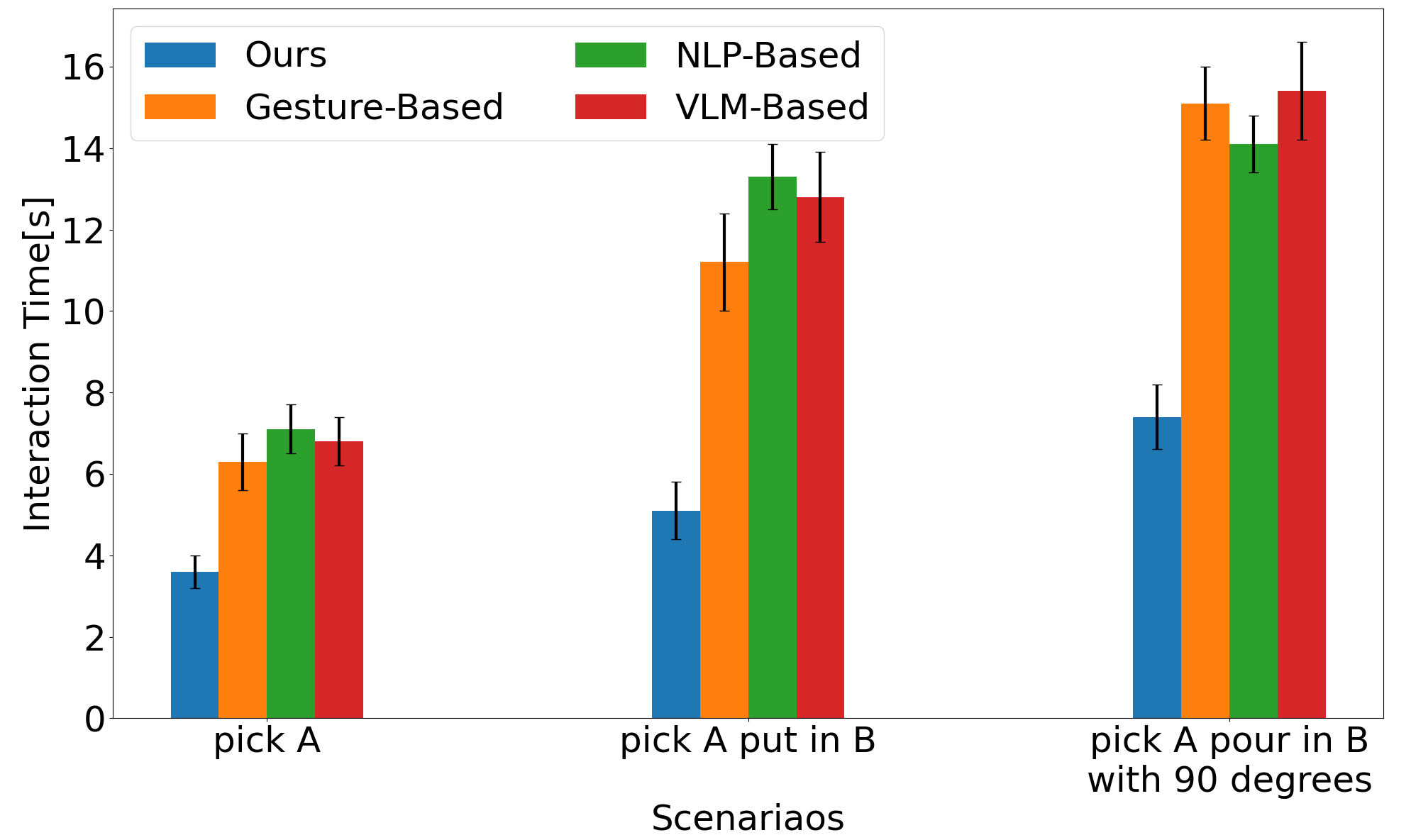}
      \vspace{-5pt}
      \caption{ Compared to other baseline HRI, our system required less time to input the same commands.}
      \label{episode}
      \vspace{-15pt}
   \end{figure}

\subsection{Comparison of Efficiency and Intuitiveness} 
In the experimental setup, we used two cups of different colors, two bowls of different colors and a plate as manipulated objects. To evaluate the time required for user interaction, we asked participants to perform the same commands using different HRI methods. These tasks included picking up the assigned cup, picking up the assigned cup and placing it on the plate, and picking up the assigned cup and pouring it into the assigned bowl at a 90-degree angle.
For the \textcolor{black}{in-house} HRI experiment, we recruited 27 participants from the local university, including 6 elderly. All participants were able to speak English and received verbal briefings on the interaction method, but did not undergo any training or trial.  Fig. \ref{episode} shows the results of the experiment. The findings indicated that our system required 50.6\% less time compared to hand gesture-based HRI, 53.3\% less time compared to language-based HRI, and 54\% less time than VLM-based HRI. \textcolor{black}{
This experiment demonstrates that, rather than relying on complex gestures to convey action intentions or ambiguous language (including interactive dialogue) to specify object intentions, our system utilizes simple multimodal commands. This approach significantly improves the efficiency of human-robot interaction (HRI) compared to the three baseline methods. Additionally, the system's wide field of view enables users to interact from a greater range of positions than is possible with the Leap Motion sensor.} Overall, our proposed system is more efficient in terms of interaction in a cluttered environment.

\subsection{\textcolor{black}{Comparison of Accuracy}} 
 \textcolor{black}{
To evaluate the accuracy of action sequences in our proposed method, we designed a set of experiments using various sequences of events and compared them against several baseline methods. Accuracy was defined as the proportion of successfully executed commands ($N_{\text{executed}}$) to the total number of trials ($N_{\text{trials}}$), calculated as $ \text{Accuracy} = (N_{\text{executed}} / N_{\text{trials}})* 100\% $.  
}

\begin{figure}[t]
      \centering
      \includegraphics[width=0.5\textwidth]{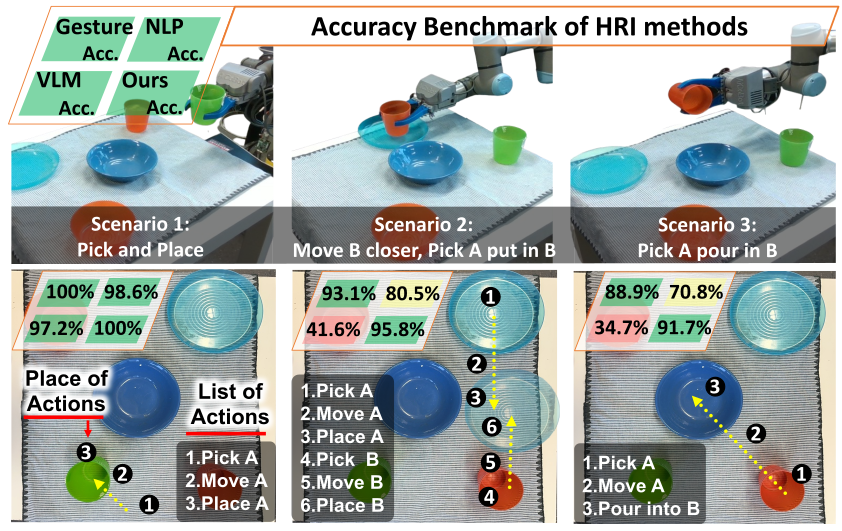}
       \vspace{-5pt}
      \caption{
Accuracy evaluation for various tasks across different approaches..}
      \label{accuracy}
       \vspace{-15pt}
\end{figure}

 \textcolor{black}{
The actions were categorized into simple commands (e.g., pick and place), and causality commands (tasks involving cause-and-effect), andsequential commands (multi-step actions). The graphical results are presented in Fig. \ref{accuracy} Gesture-based methods often outperformed language-based methods for precise object referencing but struggled with more complex action sequences. Methods relying on VLM or NLP required highly descriptive input, making them less accurate and effective for tasks involving similar objects.  
}

%Our multimodal HRI approach eliminates the need for complex gestures and reduces ambiguity in language-based instructions. With the contextual understanding capabilities of the LLM, the system corrects misrecognized voice commands, accurately interprets user intent, and generates precise action sequences. Overall, our method combines the strengths of descriptive language and intuitive commands, demonstrating high accuracy and effectiveness across diverse scenarios.
Our multimodal HRI approach simplifies interaction by eliminating complex gestures and reducing language ambiguity. Leveraging LLMs, it corrects misrecognized commands, interprets intent, and generates precise actions. This method effectively combines descriptive language with intuitive commands, ensuring high accuracy across diverse scenarios.

\begin{figure*}[t]
%%\vspace{-10pt}
\centering
\includegraphics[width=1\textwidth]{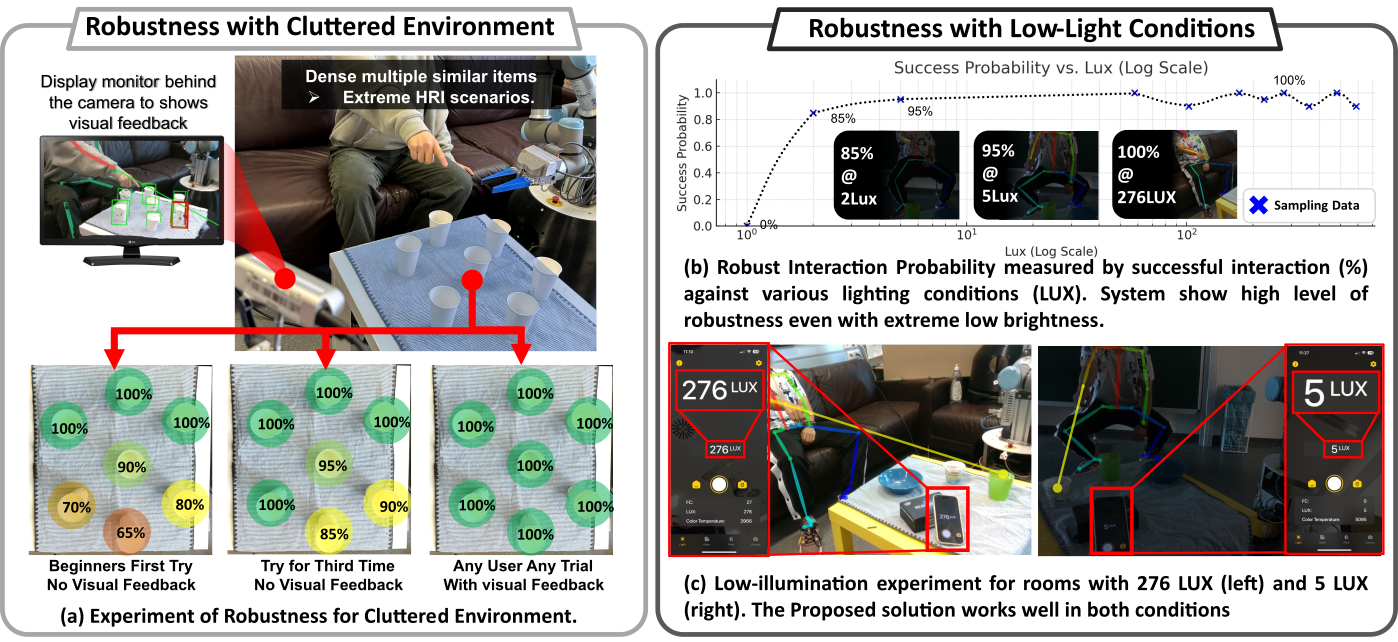}
\vspace{-5pt}
\caption{ System robustness evaluation under varying real-world constraints. Lighting affects all baseline methods equally. } 
\label{fig1:visualfeedback}
\vspace{-15pt}
\end{figure*}

%\vspace{-15pt}

\subsection{\textcolor{black}{Comparison of Robustness}}
\textcolor{black}{
To evaluate the robustness of our solution, we conducted two separate experiments. The first experiment tested robustness in a cluttered environment, while the second evaluated performance under varying lighting conditions. In both cases, we are trying to simulate complex real-world challenges. Robustness, in this context, is defined as the ability of system to correctly interpret and execute human intentions under challenging conditions. It is calculated as the proportion of correctly detected intents ($N_{\text{correct}}$) to the total number of trials ($N_{\text{total}}$), given by $ \text{Robustness} = (N_{\text{correct}} / N_{\text{total} })* 100\% $.
}

\textcolor{black}{
\textbf{Robustness Testing Cluttered Environment:} This experiment assessed the robustness of the HRI system in distinguishing between multiple similar objects, a common challenge in real-world scenarios. Six cups, separated by 25 cm, were arranged on a table to evaluate the accuracy intent detection under cluttered conditions, as shown in Fig. \ref{fig1:visualfeedback}(a).
}

\textcolor{black}{
To further test the robustness of the system, we conducted experiments with 27 participants divided into two groups. All participants received verbal instructions on using the NMM-HRI system. The first group completed three trials without visual feedback, demonstrating improved accuracy after brief learning sessions. The second group operated with visual feedback, achieving consistently high accuracy, highlighting the system's reliability in providing clear and intuitive interactions. The results reveal that the system’s robustness is influenced by the precision of human skeleton detection and the quality of the point cloud generated by the depth camera. These findings underscore the system’s capacity for reliable performance in complex, cluttered environments.
}

\textcolor{black}{
\textbf{Robustness Testing in Low-Light Conditions:}
To assess the system’s robustness under low-light conditions, we conducted tests across a lighting range of 1 to 600 lux, as shown in Fig. \ref{fig1:visualfeedback}(b) and Fig. \ref{fig1:visualfeedback}(c).  The data indicate that the system’s HRI accuracy in low-light conditions is comparable to its performance under normal lighting levels.}

\textcolor{black}{
As illustrated in \ref{fig1:visualfeedback}(b), the system effectively captures human intention when the light is above 1 Lux. These findings confirm the adaptability of the system for most typical lighting conditions. However, in extreme low-light scenarios (e.g., At 1 lux or complete darkness), vision-based methods \cite{wang2017heterogeneous} fail unless thermal imaging is used. While thermal imaging could address this issue, its high cost makes it impractical for elderly care applications.
}

% A key concern is the robustness of the system under low light conditions. To address this, we conducted tests across a range of lighting levels, from 1 to 600 lux, with the results shown in Fig. \ref{fig1:visualfeedback}(b) and Fig. \ref{fig1:visualfeedback}(c). The data demonstrate that the system's HRI accuracy under low-light  conditions is comparable to its performance under normal lighting levels.

% \textcolor{black}{Fig. \ref{fig1:visualfeedback}(b)} further illustrates that at lighting levels above 1 lux, the proposed system effectively captures human intent, and audio performance remains stable. These findings indicate that the system is suitable for most normal lighting conditions. However, in extreme low-light scenarios (e.g., less than 1 lux or complete darkness), vision-based methods become ineffective unless thermal imaging is employed. While thermal imaging could address this limitation, its high cost makes it impractical for current elderly care applications.

\begin{figure}[t]
\centering
\includegraphics[width=0.45\textwidth]{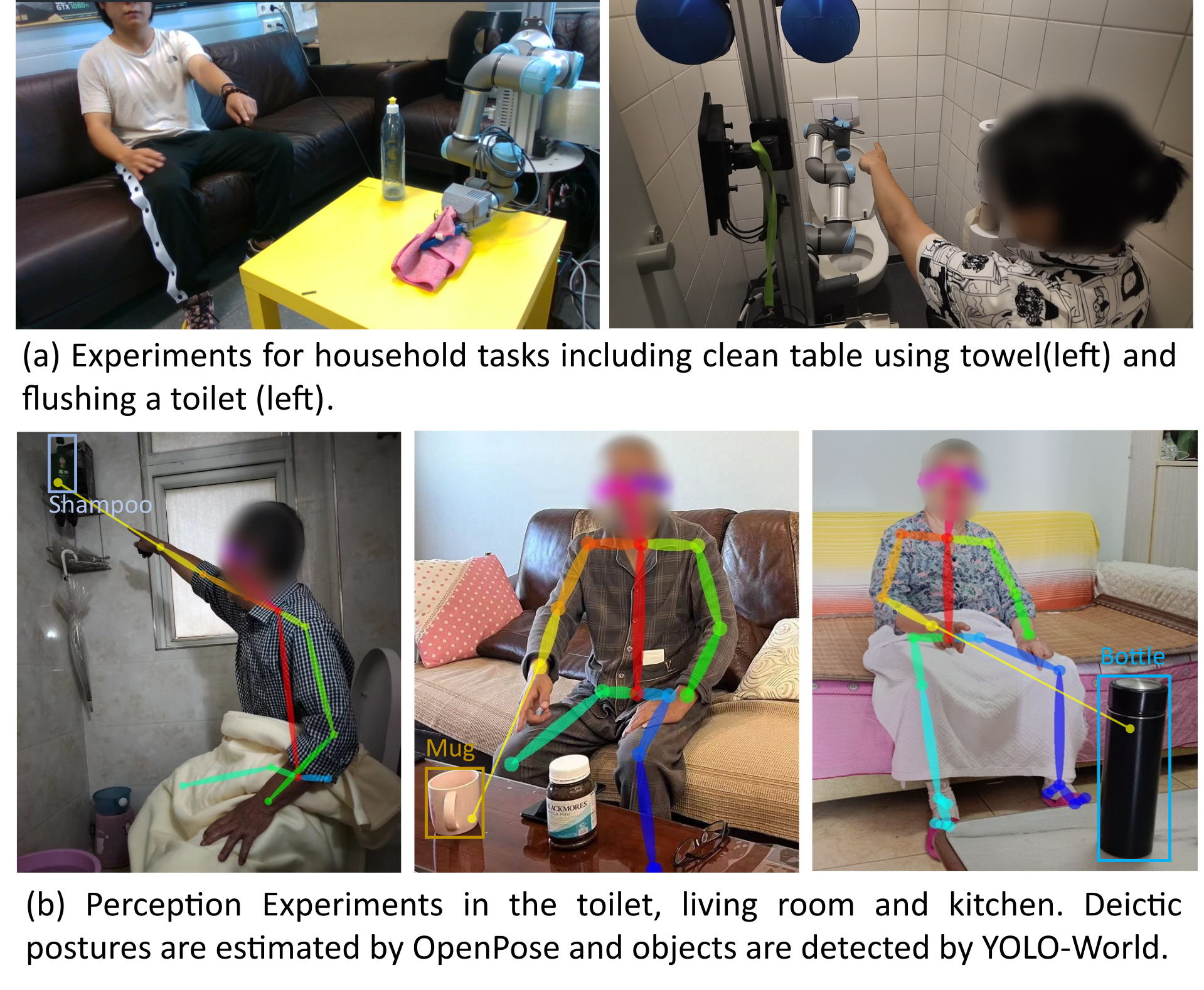}
\vspace{-8pt}
\caption{\textcolor{black}{Experiments on adverse tasks and evaluations with diverse elderly participants and environments.}} 
\label{fig1:adverse_tests}
\vspace{-15pt}
\end{figure}

\subsection{Real-world Action and Perception Trials}
\textcolor{black}{
To evaluate the versatility of the system in real-world scenarios, we conducted a series of action and perception field trials. Beyond generating action sequences for intuitive single tasks (e.g., picking up an object at location A), as shown in Fig. \ref{fig1:adverse_tests}(a), the LLM was tested on complex household tasks requiring higher-level reasoning. For example, when tasked with clearing a table using a towel, the LLM successfully generated action sequences, including locating the towel, picking it up, moving the end-effector to the table, and performing a wiping motion to clean the surface. This showcases the advanced reasoning capabilities of LLM in robotic applications.}

\textcolor{black}{
Additionally, the system was tested in diverse environments common to elderly care centers and homes, demonstrating robust performance even when the lower body of the user was obscured. In these scenarios, the system reliably fetched daily objects on demand, as shown in Fig. \ref{fig1:adverse_tests}(b). These trials highlight the system's adaptability to varying environments and its practical usability in real-world settings.}

% \subsection{Diverse Real-world Action Trials } 
% In addition to using the LLM to generate action sequences for intuitive single actions (e.g., picking up an object at location A), as shown in Fig. \ref{fig1:lowlight}(b), we also conducted experiments involving real-world household tasks requiring more complex human intentions. For instance, when tasked with clearing the table using a towel, the LLM first located the towel and then generated action sequences that included picking it up, moving the end-effector to the table, and performing a wiping motion to clean the surface. This demonstrates the LLM's complex reasoning capability with robots. 

% \subsection{Diverse Real-world Environment Perception Tests} 
% Additionally, we tested the system in various common scenarios at elderly care centers and homes, demonstrating its ability to operate effectively in different environments, even when the user's lower body is obscured. The system successfully fetched various daily objects on demand, as shown in Fig. \ref{fig1:adverse_tests}(a) and  Fig. \ref{fig1:adverse_tests}(b). 

% \begin{figure}[b]
% \centering
% \includegraphics[width=0.45\textwidth]{image/perceptionhome.png}
% \caption{\footnotesize  \textcolor{black}{Perception Experiments in the toilet (left), living room (middle), and kitchen (right). Deictic postures are estimated by OpenPose \cite{c12} and objects are detected by YOLO-World \cite{cheng2024yolo}.}} 
% \label{fig1:perceptionhome}
% %\vspace{-5pt}
% \end{figure}

\section{LIMITATION AND FUTURE WORKS}
%The proposed method has been tested in labs, homes, and elderly care centers with participants of various ages. However, most participants were fluent in English, which introduces language bias. To ensure accessibility for elderly or vulnerable patients, further testing is needed in hospitals and with speakers of other languages. While regulatory restrictions have delayed hospital trials, collaborating with healthcare professionals to conduct these tests is our key goal for the future.
The method was tested in labs, homes, and elderly care centers with diverse participants, though most were fluent in English, introducing language bias. Further testing in hospitals and with non-English speakers is needed for accessibility. Regulatory delays have impacted hospital trials, but collaboration with healthcare professionals remains a priority.

%One key limitation of NMM-HRI is its performance in extremely low-light conditions, where posture detection may be compromised and the robot cannot detect objects in complete darkness. While thermal sensors could mitigate this issue, their high cost makes them impractical. Therefore, designing specialized auxiliary sensors may be necessary to enhance performance in such environments.
NMM-HRI struggles in extreme low-light, limiting posture detection and object recognition in darkness. While thermal sensors could help, their high cost is impractical. Developing affordable specialized auxiliary sensors may be necessary.

% Our proposed method progressively improves the object detection system, currently leveraging the open-vocabulary model YOLO-World, which is highly robust for general object recognition. However, YOLO-World exhibits a significant bias toward common knowledge, resulting in near-zero accuracy for medical-specific items. To address this, a potential solution involves creating a database that enables nursing staff to update the system through verbal object identification and online domain adaptation. Exploring this approach will be a key focus in future work.
Our method enhances object detection using YOLO-World, a robust open-vocabulary model. However, its bias toward common knowledge limits accuracy for medical items \cite{lai2025nvp}. A potential solution is a database allowing nursing staff to update the system via verbal identification and online adaptation. Future work will explore this approach \cite{lai2025fam}.

% \subsection{Scenarios Evaluation}

% A total of five users with no prior experience were introduced to our parallel multimodal HRI method.
% These novice users are asked to accomplish complex scenarios through the multi-step tasks defined in sec. \ref{environment}.  Fig \ref{time} illustrates the duration of interaction and the processing for each scenario. Processing time denotes the time taken to generate action sequences through LLM. 
% Most of the errors occurred due to the misdetections of the object.
% Fluctuations in the 3D point cloud can lead to errors in coordinate calculation.
% Occlusion of the body may result in objects not being detected.
% Some voice commands might not be heard, might be recognized incorrectly, or the system might experience crashes. Due to the careful design of the prompts, the action sequences generated by GPT4 did not experience errors in the experiments.
% \begin{figure}[thpb]
%      \centering
%      \includegraphics[width=0.3\textwidth]{image/completion.png}
%      \caption{Duration of interaction and the processing for each scenario.}
%      \label{time}
%   \end{figure}

%%\vspace{-20pt}

%%\vspace{-3pt}
\section{CONCLUSIONS}
%%\vspace{-3pt}
%In this work, We proposed a system that can handle parallel multimodal inputs for HRI while handling different verbal feature types and deictic posture and combining its information to generate action sequences and execution through LLM.
In this work, we introduced a system that handles parallel multimodal inputs for HRI, accommodating diverse verbal features and dynamic postures, and integrates this information to generate action sequences executed through an LLM. Our system demonstrates excellent effectiveness, accuracy, and robustness under various conditions. We will make our code publicly available for the benefit of the community.

%Deictic posture test on users gave us information that visual feedback is a crucial component of such a system.
%Compared to previous state-of-the-art gesture-based HRI systems, our system is more intuitive, obviating the need to learn complex gestures. The average interaction time for identical simple commands can be reduced by 28.6\%. For complex environments, the interaction time can be reduced by 78.4\%.
%From the scenario test, we can see that users are able to fully grasp the use of the system with a simple explanation. 

%However, the current system is only designed for English-speaking individuals and only works with no hearing or speaking disabilities. In the future, we aim to overcome these limitations by fully incorporating gestures and other language models into the system to make it more versatile.

%In the future, we aim to refine voice commands to enable robots to understand user intentions more naturally rather than relying solely on keywords. Additionally, we plan to explore the training of efficient end-to-end small language models to reduce the time required to generate action sequences.

\addtolength{\textheight}{0cm}   % This command serves to balance the column lengths

%%%%%%%%%%%%%%%%%%%%%%%%%%%%%%%%%%%%%%%%%%%%%%%%%%%%%%%%%%%%%%%%%%%%%%%%%%%%%%%%

%%%%%%%%%%%%%%%%%%%%%%%%%%%%%%%%%%%%%%%%%%%%%%%%%%%%%%%%%%%%%%%%%%%%%%%%%%%%%%%%

\bibliographystyle{./IEEEtran}
\bibliography{./IEEEabrv,./IEEEexample}

\end{document}